\documentclass[runningheads]{llncs}
\usepackage{graphicx}
\usepackage{amsmath,amsfonts,bm}

\usepackage{booktabs} % For formal tables
\usepackage{paralist}
\usepackage{CJKutf8}
\usepackage{multirow}
\usepackage{color}

\def \a {\mathbf{a}}
\def \b {\mathbf{b}}
\def \e {\mathbf{e}}
\def \v {\mathbf{v}}
\def \p {\mathbf{p}}

\def \W {\mathbf{W}}

\def \R {\mathbb{R}}
\def \I {\mathcal{I}}

\def \C {\mathcal{C}}
\def \D {\mathcal{D}}
\def \E {\mathcal{E}}
\def \L {\mathcal{L}}
\def \CR {\mathcal{R}}

\def \KG {\mathcal{KG}}
\def \Q {\mathcal{Q}}
\def \S {\mathcal{S}}
\def \inOf {\mathsf{instanceOf}}
\def \sCOf {\mathsf{subClassOf}}

\begin{document}
\title{KGSynNet: A Novel Entity Synonyms Discovery Framework with Knowledge Graph}
\titlerunning{KGSynNet}

\author{Yiying Yang\inst{1}\textsuperscript{\textsection} \and
Xi Yin\inst{1}\textsuperscript{\textsection} \and
Haiqin Yang\inst{1}$^*$ \and
Xingjian Fei\inst{1} \and
Hao Peng\inst{2}$^*$ \and\\
Kaijie Zhou\inst{1} \and
Kunfeng Lai\inst{1} \and
Jianping Shen\inst{1}}

% %
\authorrunning{Yang and Yin et al.}
\institute{Ping An Life Insurance Company of China, Ltd., Shenzhen, China\\
\email{\{yangyiying283, yinxi445, feixingjian568, zhoukaijie002, laikunfeng597,\\shenjianping324\}@pingan.com.cn; $^*$hqyang@ieee.org}\\
\and BDBC, Beihang University, Beijing, China\\
\email{$^*$penghao@act.buaa.edu.cn}}


%
\maketitle              % typeset the header of the contribution
\begingroup\renewcommand\thefootnote{\textsection}
\footnotetext{Equal contribution.    $^*$~Corresponding authors.}
\endgroup
%\begingroup\renewcommand\thefootnote{\textsection2}
%\footnotetext{Authors contribute equally.}
%\endgroup

\begin{abstract}
Entity synonyms discovery is crucial for entity-leveraging applications.  However, existing studies suffer from several critical issues: (1) the input mentions may be out-of-vocabulary (OOV) and may come from a different semantic space of the entities; (2) the connection between mentions and entities may be hidden and cannot be established by surface matching; and (3) some entities rarely appear due to the long-tail effect.  To tackle these challenges, we facilitate knowledge graphs and propose a novel entity synonyms discovery framework, named \emph{KGSynNet}. Specifically, we pre-train subword embeddings for mentions and entities using a large-scale domain-specific corpus while learning the knowledge embeddings of entities via a joint TransC-TransE model.  More importantly, to obtain a comprehensive representation of entities, we employ a specifically designed \emph{fusion gate} to adaptively absorb the entities' knowledge information into their semantic features.  We conduct extensive experiments to demonstrate the effectiveness of our \emph{KGSynNet} in leveraging the knowledge graph. The experimental results show that the \emph{KGSynNet} improves the state-of-the-art methods by 14.7\% in terms of hits@3 in the offline evaluation and outperforms the BERT model by 8.3\% in the positive feedback rate of an online A/B test on the entity linking module of a question answering system. 

\keywords{Entity synonyms discovery  \and Knowledge graph } %  \and Knowledge fusion \and Fusion gate
\end{abstract}
%s

%\input{samplebody-conf}

\section{Introduction}
%With the prevalence of online health question-answering services, people who need healthcare consultation about certain topics, such as diseases, symptoms, and treatments, would like to search the information by themselves over the online platforms~\cite{DBLP:journals/chb/Mano14a,DBLP:journals/jbi/HuZYCZ17}.  The demand of such services becomes much higher when they are connected to both personal health and finance.  For instance, in a professional health insurance app, users may ask whether they can purchase a certain insurance product when they have specific disease symptoms: 
Entity synonyms discovery is crucial for many entity-leveraging downstream applications such as entity linking, information retrieval, and question answering (QA)~\cite{DBLP:conf/cnlp/MondalPSGPBG19,DBLP:conf/aaai/WangKMYTACFMMW19}.  For example, in a QA system, a user may interact with a chatbot as follows: 

\emph{User query:} Am I qualified for the new insurance policy as I suffer from \textbf{skin relaxation} recently?

\emph{System reply:} Unfortunately, based on the policy, you may fall into the terms of \textbf{Ehlers-Danlos}, which may exclude your protection.  Please contact our agents for more details.  

%  "Cutis" has the same meaning as the body part attribute "skin", thus we recall three candidates marked in the light blue boxes. On the other hand, "hyperelastica" has similar meaning with the symptom attribute "increased skin elasticity". Therefore, the true synonym entity "Ehlers-Danglos" is recognized out of the three candidates.
\begin{figure}[t]
  \centering
  \includegraphics[width=0.8\linewidth]{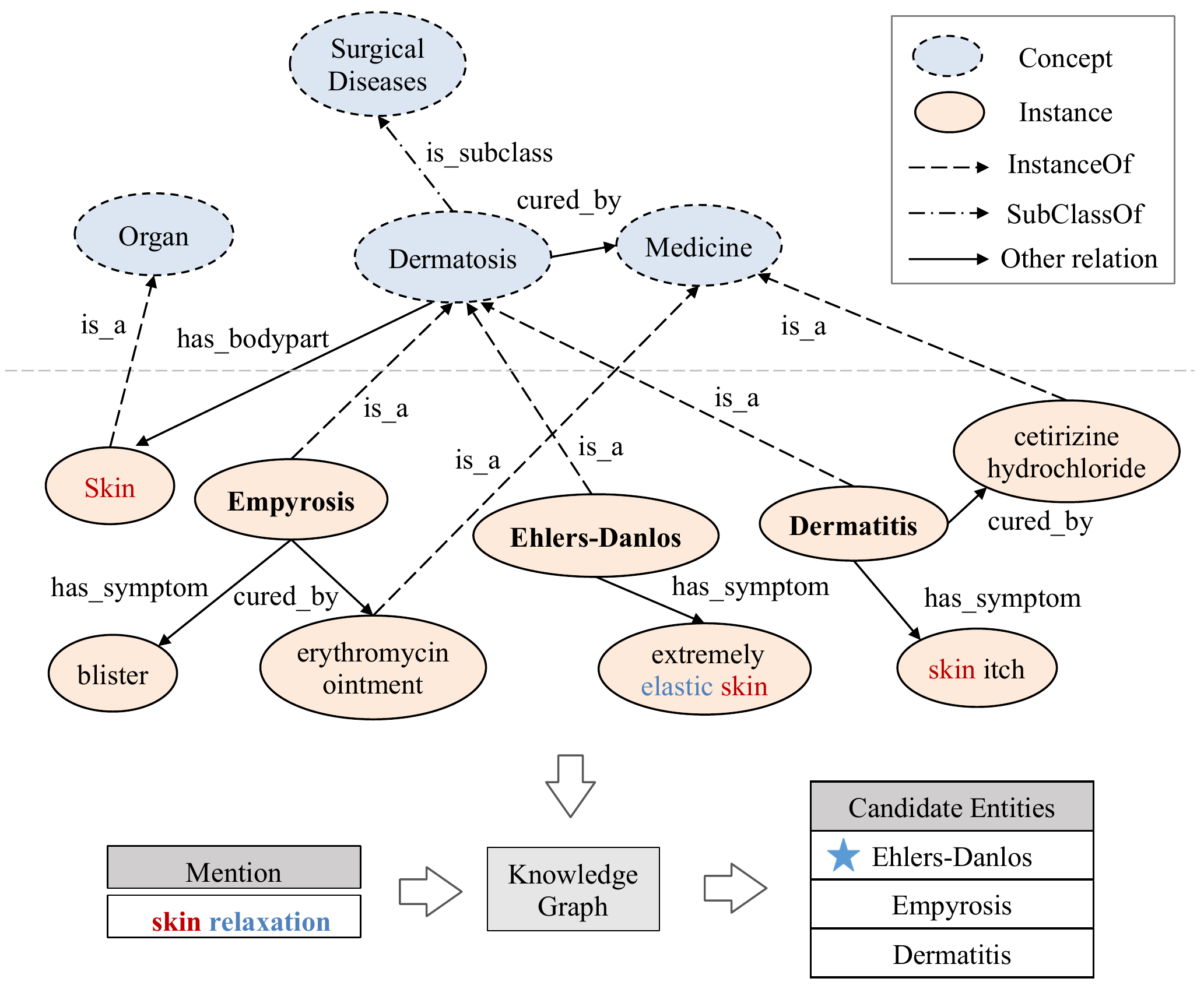}
  \caption{An illustration of linking the synonymous entity of the mention ``skin relaxation" to ``Ehlers-Danlos" with the help of an external knowledge graph. } 
  %\Description{overview}
  \label{fig:figure1}
\end{figure}

In this case, we can correctly answer the user's query only linking the mention of ``skin relaxation" to the entity, ``Ehlers-Danlos".  This is equivent to the entity synonyms discovery task, i.e., automatically identifying the synonymous entities for a given mention or normalizing an informal mention of an entity to its standard form~\cite{DBLP:conf/aaai/DoganL12,DBLP:conf/ijcai/WangCZ15}.

In the literature, various methods, such as DNorm~\cite{DBLP:journals/bioinformatics/LeamanDL13}, JACCARD-based methods~\cite{DBLP:conf/edbt/WangLLZ19}, and embedding-based methods~\cite{DBLP:journals/bmcbi/ChoCL17,DBLP:conf/kdd/FeiTL19}, have been proposed to solve this task.  They usually rely on matching of syntactic string~\cite{DBLP:conf/aaai/DoganL12,DBLP:conf/edbt/WangLLZ19} or lexical embeddings~\cite{DBLP:journals/bmcbi/ChoCL17,DBLP:conf/kdd/FeiTL19,DBLP:conf/acl/SungJLK20} to build the connections.  Existing methods suffer from the following critical issues: (1) the input mentions and the entities are often out-of-vocabulary (OOV) and lie in different semantic spaces since they may come from different sources; (2) the connection between mentions and entities may be hidden and cannot be established by surface matching because they scarcely appear 
together; and (3) some entities rarely appear in the training data due to the long-tail effect.

% distinguishing concepts from instances  in robustly modeling relations in entities the representations of mentions and entities may lie in different semantic spaces because mentions may come from spoken language while entities may be written in formal languages
To tackle these challenges, we facilitate knowledge graphs and propose a novel entity synonyms discovery framework, named \emph{KGSynNet}.  Our \emph{KGSynNet} resolves the OOV issue by pre-training the subword embeddings of mentions and entities using a domain-specific corpus.  Moreover, we develop a novel TransC-TransE model to jointly learn the knowledge embeddings of entities by exploiting the advantages of both TransC~\cite{DBLP:conf/emnlp/LvHLL18} in distinguishing concepts from instances and TransE~\cite{DBLP:conf/nips/BordesUGWY13} in robustly modeling various relations between entities.  Moreover, a \emph{fusion gate} is specifically-designed to adaptively absorb the knowledge embeddings of entities into their semantic features.  As illustrated in Fig.~\ref{fig:figure1}, our \emph{KGSynNet} can discover the symptom of ``extremely elastic skin" in the entity of ``Ehler-Danlos" and link the mention of ``skin relaxation" to it.   %a medical knowledge graph has been exploited to correctly linking the mention ``skin relaxation" to its synonymous entity ``Ehler-Danlos".  Though ``skin relaxation" exhibits neither morphological nor semantic similarity to ``Ehler-Danlos", it links to ``Ehler-Danlos" because of its symptom, ``extremely elastic skin", which is semantically closer to ``skin relaxation" than the symptom ``skin itch" in ``Dermatitis".  Obviously, this procedure relies on effectively absorbing both the knowledge and semantic features of entities. 
%{Moreover, to make more entities linkable to the mentions, we introduce external knowledge and adaptively integrate the knowledge embeddings of entities into their semantic features via a specifically-designed \emph{fusion gate}.    As illustrated in Fig.~\ref{fig:figure1}, a medical knowledge graph has been exploited to correctly linking the mention ``skin relaxation" to its synonymous entity ``Ehler-Danlos".  Though ``skin relaxation" exhibits neither morphological nor semantic similarity to ``Ehler-Danlos", it links to ``Ehler-Danlos" because of its symptom, ``extremely elastic skin", which is semantically closer to ``skin relaxation" than the symptom ``skin itch" in ``Dermatitis".  Obviously, this procedure relies on effectively absorbing both the knowledge and semantic features of entities. }  and maps their embeddings into the same semantic space to tackle the issue of data from different sources

%\if 0
In summary, our work consists of the following contributions:
\begin{compactitem}[--]
\item We study the task of automatic entity synonyms discovery, a significant task for entity-leveraging applications, and propose a novel neural network architecture, namely \emph{KGSynNet}, to tackle it.  
\item Our proposed \emph{KGSynNet} learns the pre-trained embeddings of mentions and entities from a domain-specific corpus to resolve the OOV issue.  Moreover, our model harnesses the external knowledge graph by first encoding the knowledge representations of entities via a newly proposed TransC-TransE model. Further, we adaptively incorporate the knowledge embeddings of entities into their semantic counterparts by a specifically-designed \emph{fusion gate}.  %The similarity of the encoded semantic features of mentions and the fused features of entities is then computed and optimized to seek the model parameters.
%\item We build a large-scale Chinese health insurance related knowledge graph, which consists of entities and relations in the categories of insurance products, occupation, and medicine. Moreover, we construct a massive medical disease dataset consisting of well-annotated mention-entity synonymous pairs for the experimental evaluation.
\item We conduct extensive experiments to demonstrate the effectiveness of our proposed \emph{KGSynNet} framework while providing detailed case studies and errors analysis.  Our model significantly improves the state-of-the-art methods by 14.7\% in terms of the offline hits@3 and outperforms the BERT model by 8.3\% in the online positive feedback rate.
\end{compactitem}

\begin{CJK*}{UTF8}{gbsn}
\section{Related Work}
\label{relatedwork}
Based on how the information is employed, existing methods can be divided into the following three lines: 
\begin{compactitem}[--]
\item The first line of research focuses on capturing the surface morphological features of sub-words in mentions and entities~\cite{DBLP:conf/aaai/DoganL12,DBLP:conf/acl/DSouzaN15,DBLP:conf/edbt/WangLLZ19}.  They usually utilize lexical similarity patterns and the synonym rules to find the synonymous entities of mentions.  Although these methods are able to achieve high performance when the given mentions and entities come from the same semantic space, they fail to handle terms with semantic similarity but morphological difference.

% the relation constrained model~\cite{DBLP:conf/acl/YuD14} and the hierarchical multi-task word embedding method~\cite{DBLP:conf/kdd/FeiTL19} include
\item The second line of research tries to learn semantic embeddings of words or sub-words to discover the synonymous entities of mentions~\cite{DBLP:journals/bmcbi/ChoCL17,DBLP:conf/naacl/FaruquiDJDHS15,DBLP:conf/kdd/FeiTL19,DBLP:journals/bmcbi/LiCTWXWH17,DBLP:conf/cnlp/MondalPSGPBG19}.  For example, the term-term synonymous relation has been included to train the word embeddings~\cite{DBLP:conf/kdd/FeiTL19}.  More heuristic rule-based string features are expanded to learn word embeddings to extract medical synonyms~\cite{DBLP:conf/ijcai/WangCZ15}.  These methods employ semantic embeddings pretrained from massive text corpora and improve the discovery task in a large margin compared to the direct string matching methods.  However, they perform poorly when the terms rarely appear in the corpora but reside in external knowledge bases.
\item 
The third line of research aims to incorporate external knowledge from either the unstructured term-term co-occurrence graph or the structured knowledge graph.  For example, Wang et al.~\cite{DBLP:conf/kdd/WangYMHLS19} utilizes both semantic word embeddings and a term-term co-occurrence graph extracted from unstructured text corpora to discover synonyms on privacy-aware clinical data. Jiang et al. \cite{DBLP:conf/icdm/JiangLWCLWA13} applies the path inference method over knowledge graphs.  More powerful methods, such as  SynSetMine~\cite{DBLP:conf/aaai/ShenLRVSH19}, SA-ESF~\cite{DBLP:journals/symmetry/HuTZGX19}, and the  contextualized method~\cite{DBLP:journals/jamia/SchumacherD19}, have been proposed to leverages the synonym of entities in knowledge graphs or the knowledge representations.  They ignore other relations among entities, e.g., the hypernym-hyponym relations, and lack a unified way to absorb the information.  This motivates our further exploration in this work.

% but yields a lower recall than knowledge embedding based methods, such as  SynSetMine~\cite{DBLP:conf/aaai/ShenLRVSH19}, SA-ESF~\cite{DBLP:journals/symmetry/HuTZGX19}, and the  contextualized method~\cite{DBLP:journals/jamia/SchumacherD19}.  However, SynSetMine~\cite{DBLP:conf/aaai/ShenLRVSH19} only leverages the synonym type of knowledge in the graphs and ignores other rich relations among entities (e.g., the hypernym-hyponym semantic relations), while SA-ESF~\cite{DBLP:journals/symmetry/HuTZGX19} and contextualized representation based methods~\cite{DBLP:journals/jamia/SchumacherD19} facilitate the learned knowledge representations by standard concatenation with semantic features, which yields sub-optimal performance.  {Therefore, it is crucial to investigate how to effectively learn the knowledge embedding of entities by sufficiently absorbing the relevant information in the knowledge graph and efficaciously incorporate the knowledge information into the semantic features. }
\end{compactitem}

\section{Methodology}
Here, we present the task and the main modules of our \emph{KGSynNet} accordingly.

\if 0
\subsection{Notation and Task Definition}
We first define the mathematical notation for the sake of consistence in the whole paper.  Bold capital letters, e.g., $\W$, indicate matrices.  Bold small letters, e.g., $\v$, indicate vectors.  Letters in calligraphic or blackboard bold fonts, e.g., $\R$ and $\S$, indicate sets, where $\R^n$ denotes an $n$-dimensional real space.  $m\in\R_+$ denotes a non-negative real value.  For a little abuse of notations, a common letter, e.g., $e$, denotes a function or a string.  $|\cdot|$ denotes the size of a set or a string of subwords.  
\fi 
%The operator ⊤ denotes the transpose and ⟨x, y⟩H defines the inner product of x and y in the space H. d(H) defines the dimension of the space H. X ≽ 0 denotes a matrix which is positive semidefinite. The operator ◦ defines the Hadamard or elementwise product.

%\cparagraph{Task} 
\subsection{Task definition.} 
{The task of entity synonyms discovery is to train a model to map the mention to synonymous entities as accurate as possible given a set of annotated mention-entity pairs $\Q$, a knowledge graph $\KG$, and a domain-specific corpus, {$\D$}.  The mention-entity pairs, ${\Q}= \{(q_i, t_i)\}_{i=1}^N$, record the mentions from queries and their corresponding synonymous entities, where $N$ is the number of annotated pairs, $q_i=q_{i1}\,\ldots\,q_{i|q_i|}$ denotes the $i$-th mention with $|q_i|$ subwords and $t_i=t_{i1}\,\ldots\,t_{i|t_i|}\in\E$ denotes the $i$-th entity in $\KG$ with $|t_i|$ subwords.  The knowledge graph is formalized as $\KG=\{\C, \I, \CR, \S\}$, where $\C$ and $\I$ denote the sets of concepts and instances, respectively, $\CR$ is the relation set and $\S$ is the triple set.  %$\D$ is xx
Based on the above definition, we have $\E=\C\cup\I$.  After we train the model, for a given mention, we can recommend a list of synonymous entities from the knowledge graph. }  The domain-specific corpus, {$\D$}, is used for learning the embeddings of mentions and entities. 

\begin{figure*}[htp]
  \includegraphics[width=\textwidth]{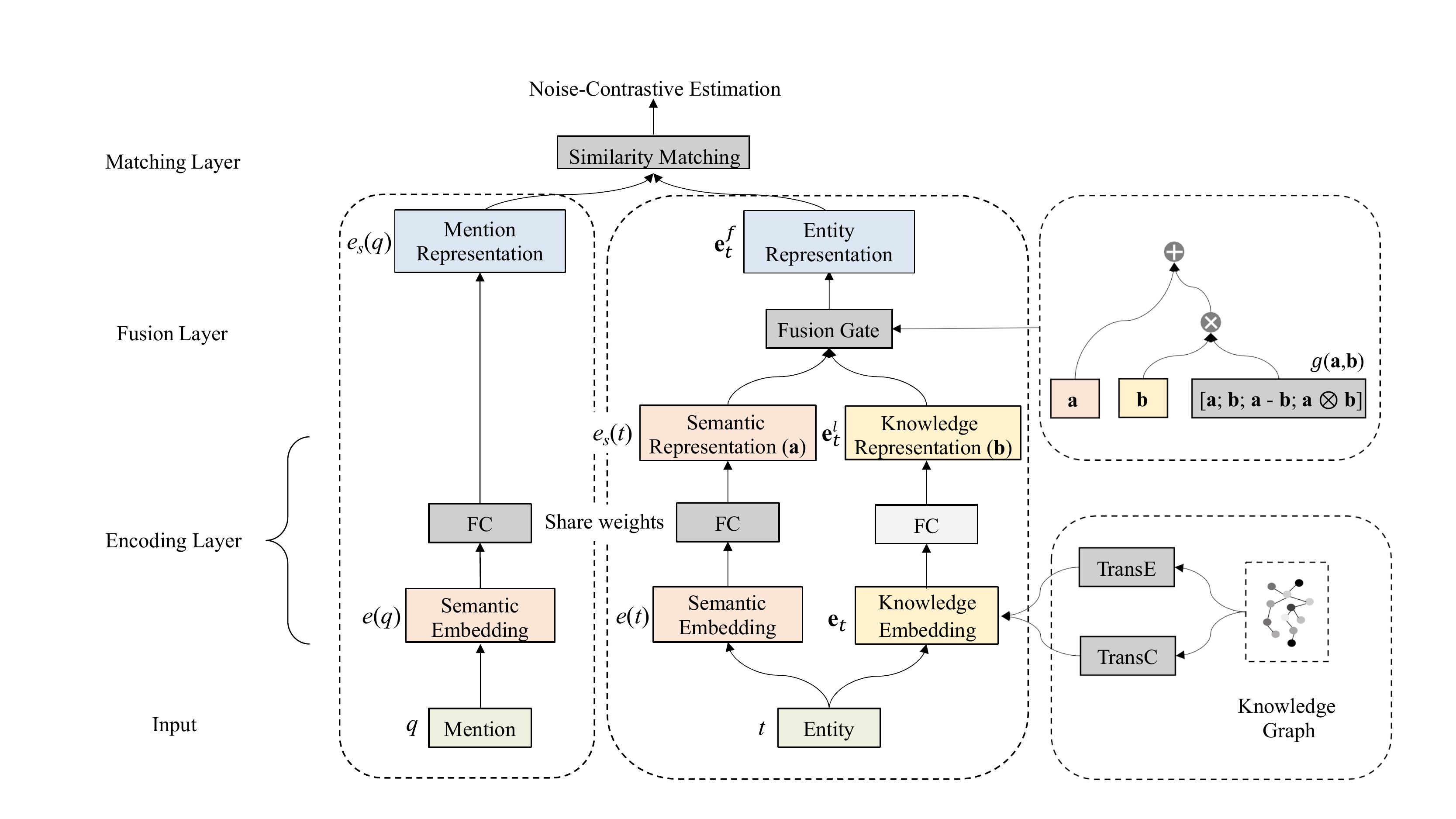}
  \caption{The architecture of our \emph{KGSynNet}.}
  % \Description{Model Architecture}: The left grid box shows the procedure of extracting semantic representations of mentions and the middle grid box shows the flow of extracting semantic and knowledge representations of entities.  It is noted that the weights of the fully-connected networks are shared to map the learned embeddings of mentions and entities in the same semantic space.  The knowledge embeddings of entities are learned from a jointly trained TransC-TransE model (shown in the bottom-right grid box) while the knowledge information is adaptively absorbed via the \emph{Fusion Gate} (shown in the upper-right grid box for the Fusion layer).   In the matching layer, similarity between mentions and entities is calculated and trained via the Noise-Contrastive Estimation (NCE).
  \label{fig:model}
\end{figure*}

%{In the following, we outline the four main modules of our KGSynNet: semantic encoder, knowledge encoder, fusion gate, and the classifier.} 
{As illustrated in Fig.~\ref{fig:model}, our proposed \emph{KGSynNet} consists of four main modules: (1) a semantic encoder module to represent mentions and entities; (2) a knowledge encoder module to represent the knowledge of entities by a jointly-learned TransC-TransE model; (3) a feature fusion module to adaptively incorporate knowledge information via a specifically designed \emph{fusion gate}; (4) a classifier with a similarity matching metric to train the entire model. } 

%\fi 

%\subsection{Semantic Encoder}\label{sec:semantic_enc} Specifically, we adopt sub-word embeddings, i.e., sub-words for Chinese words or letter tri-grams for English words~\cite{DBLP:conf/cikm/HuangHGDAH13}, due to their effectiveness in handling OOV issue.  
%\noindent {\bf Semantic encoder.} 
\subsection{Semantic Encoder}
Given a mention-entity pair, $(q, t)$, we may directly apply existing embeddings, e.g., Word2Vec~\cite{DBLP:conf/nips/MikolovSCCD13}, or BERT~\cite{DBLP:conf/naacl/DevlinCLT19}, on $q$ and $t$ to represent the semantic information of mentions and  entities.  However, it is not effective because many subwords are out-of-vocabulary (OOV), since the pre-trained embeddings are trained from corpora in general domains.  

To leverage the contextualized information of each mention and entity from $\D$, we train a set of subword-level Word2Vec embeddings from scratch on $\D$, and apply them to initialize the semantic representations of the subwords of the mentions and the entities in $\Q$.  
% That is, we map each sub-word in $q$ and $t$ via a linear mapping matrix $\W\in\R^{|\V|\times d}$ into a $d$-dimensional representation, where $|\V|$ is the vocabulary size and $d$ is the dimension of the semantic embedding. 
Then, similar to the fastText approach~\cite{DBLP:journals/tacl/BojanowskiGJM17}, we obtain the initialized semantic representations of mentions and entities by averaging their subword representations:  %m to get the corresponding semantic features for each mention and the corresponding entity and represent them as follows:  
%\if 0
\begin{equation} \label{eq:semantic_embedding}
e(q) = \frac{1}{|q|}\sum_{k=1}^{|q|}e(q_k),\quad e(t)= \frac{1}{|t|}\sum_{k=1}^{|t|}e(t_k).
\end{equation}
%\fi 
%$ and $$, respectively.         

After that, the semantic embeddings of the mentions and the entities are further fed into a two-layer fully-connected (FC) network to extract deeper semantic features.  Here, we adopt shared weights as in~\cite{DBLP:conf/acl/ChenZLWJI17} to transform the learned embedding $e(v)$ into a semantic space of $k$-dimension:
\begin{equation}
e_s(v) = \tanh(\W_2\tanh(\W_1e(v)+b_1)+b_2)\in \R^k,
%e_s(t) & = & \tanh(\W_2\tanh(\W_1e(t)+b_1)+b_2),
\end{equation}
where $v$ can be a mention or an entity. The parameters, $\W_1\in\R^{k\times d}$ and $\W_2\in\R^{k\times k}$, are the weights on the corresponding layers of the FC network.  $b_1\in\mathbb{R}^k$ and $b_2\in\mathbb{R}^k$ are the biases at the corresponding layers.

\subsection{Knowledge Encoder}
%\noindent {\bf Knowledge encoder for entities.} 
Though entities can be encoded in the semantic space as detailed above, their representations are not precise enough due to lack of the complementary information included in the knowledge graph.  %In the section, we focus on exploiting the knowledge graph to enrich the representation of the entities. 

In the knowledge graph $\KG$, the relation set $\CR$ is defined by $\CR=\{r_e, r_c\}\cup\CR_l\cup\CR_{\scriptscriptstyle\I\scriptscriptstyle\C}\cup\CR_{\scriptscriptstyle\C\scriptscriptstyle\C}$, where $r_e$ is an $\inOf$ relation, $r_c$ is a $\sCOf$ relation, $\CR_l$ is the instance-instance relation set, $\CR_{\scriptscriptstyle\I\scriptscriptstyle\C}$ is the Non-Hyponym-Hypernym (NHH) instance-concept relation set, and $\CR_{\scriptscriptstyle\C\scriptscriptstyle\C}$ is the NHH concept-concept relation set.  It is noted that different from the three kinds of relations defined in TransC~\cite{DBLP:conf/emnlp/LvHLL18}, we specifically categorize the relations into five types to differentiate the NHH relations of the instance-concept pairs from the concept-concept pairs.  Therefore, the triple set $\S$ can be divided into the following five disjoint subsets:
\begin{compactenum}[1.]
\item The $\inOf$ triple set: $\S_e=\left\{\begin{pmatrix}i, r_e, c
\end{pmatrix}_k\right\}_{k=1}^{|\S_e|}$, where $i\in\I$ is an instance, $c\in\C$ is a concept, and $r_e$ is the $\inOf$ relation.
\item The $\sCOf$ triple set: $\S_c=\left\{\begin{pmatrix}c_i, r_c, c_j\end{pmatrix}_k\right\}_{k=1}^{|\S_c|}$, where $c_i, c_j\in\C$ are concepts, $c_i$ is a sub-concept of $c_j$, and $r_c$ is the $\sCOf$ relation.
\item The instance-instance triple set: $\S_l=\left\{\begin{pmatrix}i, r_{ij}, j\end{pmatrix}_k\right\}_{k=1}^{|\S_l|}$, where $r_{ij}\in\R_l$ defines the instance-instance relation from the head instance $i$ to the tail instance $j$.
\item The NHH instance-concept triple set: $\S_{\scriptscriptstyle\I\C}=\left\{\begin{pmatrix}i, r_{ic}, c\end{pmatrix}_k\right\}_{k=1}^{|\S_{\scriptscriptstyle\I\C}|}$, where $i$ and $c$ are defined similarly as $\S_e$. $r_{ic}\in\CR_{\scriptscriptstyle\I\C}$ is an NHH instance-concept relation. % the same meaning
%$i\in\I$ is an instance, $c\in\C$ is a concept, $r_{ic}\in\CR_{\scriptscriptstyle\I\C}$ is an NHH instance-concept relation. 
\item The NHH concept-concept triple set: $\S_{\scriptscriptstyle\C\C}=\left\{\begin{pmatrix}c_i, r_{c_ic_j}, c_j\end{pmatrix}_k\right\}_{k=1}^{|\S_{\scriptscriptstyle\C\C}|}$, where $c_i, c_j\in\C$ denote two concepts, $r_{c_ic_j}\in\CR_{\scriptscriptstyle\C\C}$ is an NHH concept-concept relation. 
\end{compactenum}

We now learn the knowledge embeddings of entities.  Since TransE~\cite{DBLP:conf/nips/BordesUGWY13} is good at modeling general relations between entities while TransC~\cite{DBLP:conf/emnlp/LvHLL18} excelling in exploiting the hierarchical relations in the knowledge graph, we propose a unified model, the TransC-TransE model, to  facilitate the advantage of both models.  

Specifically, TransE represents an entity by $\v\in \R^n$, where $n$ is the size of the knowledge embedding, and defines the loss for the instance-instance triples~\cite{DBLP:conf/nips/BordesUGWY13}:
\begin{equation}
\label{eq:ee_loss}
  f_{l}(i, r_{ij}, j)=\|\v_i+\v_{r_{ij}}-\v_j\|_2^2,
\end{equation}
where $(i, r_{ij}, j)\in \S_l$ denotes a triple in the instance-instance relation set,  $\v_i$, $\v_{r_{ij}}$, and $\v_j$ denote the corresponding TransE representations.

In TransC, an instance $i$ is represented by a vector, $\v_i\in\R^n$, same as that of an entity in TransE. A concept $c$ is represented by a sphere, denoted by $(\p_c, m_c)$, where $\p_c\in\R^n$ and $m_c\in\R_+$ define the corresponding center and radius for the concept, respectively.  The corresponding losses can then be defined as follows: 
\begin{compactitem}[{--}]
\item The loss for the $\inOf$ triples~\cite{DBLP:conf/emnlp/LvHLL18}:
\begin{equation}
\label{eq:ic_loss}
  f_{e}(i,c)=\|\v_i-\p_c\|_2-m_c,\quad \forall i\in c.
\end{equation}
%where $i$ is an instance of a concept $c$, i.e., $i\in c$.
\item The loss for the $\sCOf$ triples~\cite{DBLP:conf/emnlp/LvHLL18}:
\begin{equation}\label{eq:cc_loss}
\!\!f_{c}(c_i, c_j)=\left\{
\begin{array}{l@{}l@{}r}
m_{c_i}-m_{c_j}, & {c_j} \mbox{ is a subclass of } &{c_i} \mbox{, or }c_j\subseteq c_i \\ % 
{\|{\p_{c_i}}-{\p_{c_j}}\|}_2&~+~m_{c_i}-m_{c_j}, &  \mbox{otherwise}
\end{array} \right.. 
\end{equation}
\end{compactitem}
However, the spherical representation is not precise enough to model the NHH relations.  We therefore denote the concept of $c$ by an additional node embedding, $\v_c\in\R^n$, and define the following additional loss functions:   
\begin{compactitem}[--]
\item The loss for the NHH instance-concept triples~\cite{DBLP:conf/nips/BordesUGWY13}:
\begin{equation}\label{eq:ic_nhh_loss}
  f_{\scriptscriptstyle\I\C}(i, r_{ic}, c)=\|\v_i+\v_{r_{ic}}-\v_c\|_2^2,
\end{equation}
where the triplet $(i, r_{ic}, c)\in\S_{\scriptscriptstyle\I\C}$ denotes the NHH instance-concept relation $r_{ic}$ connecting the instance $i$ to the concept $c$.
\item The loss for the NHH concept-concept triples~\cite{DBLP:conf/nips/BordesUGWY13}:
\begin{equation}\label{eq:cc_nhh_loss}
  f_{\scriptscriptstyle\C\C}(c_i, r_{c_ic_j},  c_j)=\|\v_{c_i}+\v_{r_{c_ic_j}}-\v_{c_j}\|_2^2,
\end{equation}
where the triplet $(c_i, r_{c_ic_j}, c_j)\in\S_{\scriptscriptstyle\C\C}$ denotes the NHH concept-concept relation $r_{c_ic_j}$ connecting the concept $c_i$ to the concept $c_j$.
\end{compactitem}    
% {For an instance-instance triple $(i, r_{ij}, j)\in\S_l$ in the instance-instance relation set, we can then define the corresponding loss $f_{l}(i, r_{ij}, j)$ as that in  Eq.~(\ref{eq:ee_loss}). } 

Therefore, the knowledge embeddings of entities are learned by minimizing the following objective function:
\begin{eqnarray}\nonumber 
\L_k&=& \sum_{\scriptscriptstyle(i, r_e, c)\in\S_e} f_{e}(i, c) +  \sum_{\scriptscriptstyle(c_i, r_c, c_j) \in \S_c}f_{c}(c_i, c_j) +  \sum_{\scriptscriptstyle(i, r_{ij}, j) \in \S_l}f_{l}(i, r_{ij}, j)\\ \label{eq:TransC-TransE}\!\!&\!\!+\!\!&\!\! \sum_{\scriptscriptstyle(i, r_{ic}, c)\in\S_{\scriptscriptstyle\I\C}}f_{\scriptscriptstyle\I\C}(i, r_{ic}, c) + \sum_{\scriptscriptstyle(c_i, r_{c_ic_j}, c_j)\in\S_{\scriptscriptstyle\C\C}}f_{\scriptscriptstyle\C\C}(c_i, r_{c_ic_j}, c_j).
\end{eqnarray}
It is noted that our objective differs from TransC by explicitly including both the NHH instance-concept relations and the NHH concept-concept relations.  Similarly, we apply the negative sampling strategy and the margin-based ranking loss to train the model as in~\cite{DBLP:conf/emnlp/LvHLL18}.

After training the unified TransC-TrainsE model in Eq.~(\ref{eq:TransC-TransE}), we obtain the knowledge embeddings for both instances and concepts, e.g., $\v_i$ for an instance $i$, and the representation of $(\p_c, m_c)$ and $\v_c$ for a concept $c$.  For simplicity and effectiveness, we average the center and the node embedding of a concept to yield its final knowledge embedding $\e_t$:
\begin{equation}
\label{eq:entity_knowledge_emb}
\e_t=\left\{
\begin{array}{lcl}
\v_t       &      & \forall t\in\I \\
(\p_t+\v_t)/2     &      & \forall t\in\C
\end{array} \right.. 
\end{equation}

%where $v\in \mathbb{R}^q$, $p\in \mathbb{R}^q$, $q$ is the dimension of the original entity knowledge embedding.

\if 0
\begin{figure}[htp]
  \centering
  \includegraphics[width=\linewidth]{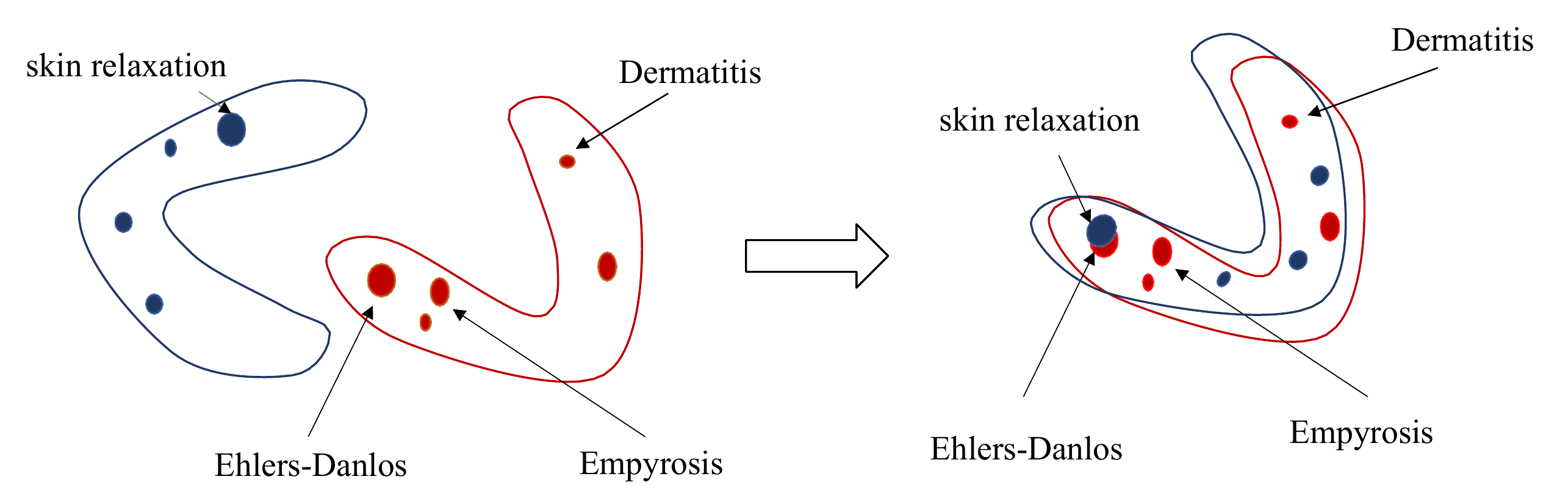}
  \caption{An illustration of aligning the mentions and entities in the same semantic space.}
  \label{fig:alignment}
\end{figure}
\fi 
%\subsection{Features Alignment and Fusion}
%Fusing Semantic and Knowledge Representations of Entities}to guarantee the semantic embeddings lie in the same semantic space as illustrated in Fig.~\ref{fig:alignment}, 

Similar to the semantic embeddings, the learned knowledge embeddings of entities obtained in Eq.~(\ref{eq:entity_knowledge_emb}) are transformed into the same $k$-dimensional semantic space by a two-layer fully connected network to yield $\e_t^l$:
\begin{equation}
\e_t^l = \tanh(\W_4(\tanh(\W_3\e_t+b_3))+b_4) \in \R^k,
\end{equation}
where $\W_3\in\mathbb{R}^{k\times q}$ and $\W_4\in\mathbb{R}^{k\times k}$ are the weights on the corresponding layers of the FC network.  $b_3\in\R^k$ and $b_4\in \mathbb{R}^k$ are the biases at the layers.  %$\sigma$ represents the activation function $tanh$, and $e_k^l \in \mathbb{R}^k$.

% As our final target is to match between mentions and entities, we need to guarantee the consistency of the semantic representations between mentions and entities. and directly feed all the semantic information of the entities to the next level without filtering set the carry gate to 1 
%\noindent {\bf Fusion gate.} 
\subsection{Fusion Gate}
A critical issue in the task is that the semantic features and the knowledge embeddings are learned separately.  To effectively integrate these two types of information, we design a fusion network, named \emph{Fusion Gate}, to adaptively absorb the transformed knowledge information $\e_t^l$ into the semantic information $e_s(t)$ for an entity $t$.   As illustrated in the upper right grid box of Fig.~\ref{fig:model}, the final representation of an entity $t$ is computed by 
\begin{equation}
    \e_t^f = e_s(t) +  e_t^l\otimes g(e_s(t), \e_t^l). \label{eq5}
\end{equation}
Here, the implementation is motivated by the highway network~\cite{DBLP:conf/nips/SrivastavaGS15}, but is different on the specific information carrying.  Here, we directly feed all the semantic information of the entities to the next level without filtering to guarantee the consistency of the semantic representations between mentions and entities.  The interaction of the semantic embeddings and knowledge embeddings of the entities is then fulfilled by the transform gate to determine the amount of knowledge incorporated into the semantic feature, defined by $g(\a, \b)$: 
\begin{equation}
g(\a, \b) = \mbox{Softmax}(\W_g[\a; \b; \a-\b; \a\otimes \b]),
\end{equation}
where $\W_g\in\mathbb{R}^{k\times4k}$ is the weight of a fully-connected network to reduce the dimension of the concatenated features.  The first two features maintain the original form while the latter two measuring the ``similarity" or ``closeness" of the two features.  This allows to compute the high-order interactions between two input vectors \cite{DBLP:conf/acl/ChenZLWJI17,DBLP:conf/acl/MouMLXZYJ16}.  Finally, the $\mbox{Softmax}$ operator is applied to determine the proportion of the flow-in knowledge.  

% \subsection{Similarity Calculation and Training Procedure}  mention-to-entity
%\noindent {\bf Classifier.} 
\subsection{Similarity Matching and Classification}
As the training data only consist of the positive  pairs, for each pair $(q_i, t_i)$, we additionally sample some negative pairs $\{(q_i, t_{i_j})\}_{j=1}^{N_i}$, where $t_{i_j}$ is sampled from other mention-to-entity pairs and $N_i$ is the number of sampled negative pairs.  Hence, we derive the objective function for the final matching: 
\begin{equation}\label{eq:loss_function}
\!\!\L_{{m}}\! =\!\sum_{i=1}^{N}-\log\left(
  \frac{\exp\left(e_s(q_i)^T\e_{t_i}^f\right)}
  {\exp\left(e_s(q_i)^T\e_{t_i}^f\right)
    + \sum_{j=1}^{N_i} \exp\left(e_s(q_i)^T \e_{t_{i_j}}^f\right)
  }\right).
\end{equation}
It is noted that each term in Eq.~(\ref{eq:loss_function}) defines the Noise-Contrastive Estimation (NCE)~\cite{DBLP:conf/aistats/GutmannH10}, which is the cross-entropy of classifying the positive pair ($q_i$, $t_i$).  After training, given a new mention $q$, we can determine the list of the candidate entities by the rank of $e_s(q_i)^T\e_{t_i}^f$.
\if 0
\begin{equation}
    e_s(q_i)^T\e_{t_i}^f.
\end{equation}
\fi

\begin{table}[htp]
\centering
\caption{Data statistics. % E\_type and R\_type denote the types of entities and relations, respectively.
}
\label{tab:statistics}
\begin{tabular}{@{}l@{~~}r@{~}r@{~}r@{$\quad$}r@{~}r@{}}
%{@{}p{2cm}@{~~}r@{~~}r@{~~}r@{~~}r@{~}r@{~}}
\hline
%\begin{tabular}[c]{@{}l@{}}\end{tabular} 
Knowledge & {All} & Insurance & Occupation & Medicine & Cross \\ 
Graph & & & & & Domain\\ \hline
\#\,Entities & 75,153 & 1,409 & 2,587 & 71,157 & 0 \\
%\begin{tabular}[c]{@{}l@{}}\# Entity \\ type\end{tabular} 
\#\,Entity\_type 
& 17 & 2 & 2 & 13 & 0 \\
\#\,Relations & 1,120,792 & 2,827 & 2,580 & 1,098,280 & 17,105 \\
%\begin{tabular}[c]{@{}l@{}}\# Relation \\ types\end{tabular} 
\#\,Relation\_type
& 20 & 2 & 2 & 13 & 4 \\ \hline
\end{tabular}
\begin{tabular}{@{}p{6.5cm}@{~~}r@{}}
\hline
\#\,Mention-entity pairs in Train/Dev/Test & 45,500/5,896/5,743 \\
\#\,Regular cases/\# Difficult cases & 5,303/440 \\ \hline
\end{tabular}
\end{table}

\section{Experiments}\label{sec:exp}
In the following, we present the curated dataset along with the associated knowledge graph, as well as the experimental details. %our constructed knowledge graph, the dataset, and our experiments with detailed analysis. 
\if 0
We conduct extensive experiments to address the following questions: 
\begin{compactitem}[--]
\item How to build a benchmark dataset with the  knowledge graph for fairly comparisons on tackling the medical entity synonymous discovery task? 
\item What is the performance of our KGSynNet in both offline and online evaluations comparing with the state-of-the-art methods and the effects of different modules?
\item What are the typical successfully discovering examples and the common errors attained by our KGSynNet?  
\end{compactitem}
{Sec.~4.1, Sec.~\ref{sec:comparison}-\ref{sec:online_eval}, and Sec.~\ref{sec:cases}-\ref{sec:errors} answer the above questions accordingly. 
}\fi

%\noindent{\bf Knowledge graph.} 
\subsection{Datasets}
\label{sec:dataset} 

%\cparagraph{Knowledge graph}  
\noindent{\bf Knowledge graph.} The existing open-source knowledge graphs~\cite{DBLP:journals/ws/BizerLKABCH09,DBLP:conf/sigmod/BollackerEPST08} cannot be used for this task, because they do not provide sufficient disease entities and relations required by the task. Therefore, we construct a specific knowledge graph ($\KG$) to verify this task. Table~\ref{tab:statistics} records the statistics of the constructed $\KG$, a heterogeneous $\KG$ with entities collected from three categories: \emph{Insurance Products}, \emph{Occupation}, and \emph{Medicine}.  In \emph{Insurance Products}, there are 1,393 insurance products and 16 concepts; while in \emph{Occupation}, there are 1,863 instances and 724 concepts obtained from the nation’s professional standards~\footnote{\scriptsize http://www.jiangmen.gov.cn/attachment/0/131/131007/2015732.pdf}.  Both \emph{Insurance Products} and \emph{Occupation} contain only two types of relations, i.e., the $\inOf$ relation and the $\sCOf$ relation.  In \emph{Medicine}, 45K disease entities and 9,124 medical concepts are extracted from three different resources: (1) raw text of insurance products’ clauses; (2) users' query logs in the app; (3) the diagnostic codes of International Classification of Diseases (ICD-10).  Furthermore, 18K other types of medical entities, such as symptom, body part, therapy, and treatment material, are extracted from some   open-source medical knowledge graphs~\footnote{\scriptsize http://openkg.cn/dataset/symptom-in-chinese; http://openkg.cn/dataset/omaha-data.}. The relation types include not only $\inOf$ and $\sCOf$, but also the instance-instance relations, the NHH concept-instance relations, and the NHH concept-concept relations, 13 types in total. %}\,\footnote{

\noindent{\bf Data.} We collect a large-scale Chinese medical corpus from 14 medical textbooks~\footnote{\scriptsize https://github.com/scienceasdf/medical-books}, 3 frequently used online medical QA forums, and some QA forums~\footnote{\scriptsize https://github.com/lrs1353281004/Chinese\_medical\_NLP}.  We also deploy a self-developed BERT-based NER tool to extract 100K disease mentions from users' query logs in the professional app.  From the extracted disease mentions and $\KG$ entities, we generate 300K candidate synonymous mention-entity pairs based on the similarity score computed by BERT.  The extracted mention-entity candidates are double-blindly labeled to obtain 57,139 high-quality disease mention-entity synonym pairs. After that, the dataset is randomly split into the sets of training, development, and test, respectively, approximately at a ratio of 8:1:1.  We further divide the test set (the {All} case group) into two groups based on the surface form similarity.  That is, a {Regular} case means that there is at least one identical subword between the mention and the entity, while the rest pairs belong to the {Difficult} case group.

% \subsection{Comparisons}\label{sec:comparison}to prevent the medical term OOV representation, mentioned in Sec.\ref{sec:dataset} input are represented by the newly trained word2vec embeddings  the mentions and the entities to  simplify computation, we apply
%\noindent{\bf Compared methods.} 
\subsection{Compared Methods}
We compare \emph{KGSynNet} with the following strong  baselines:
\begin{compactenum}[(1)]
\item JACCARD~\cite{DBLP:conf/imecs/NiwattanakulSNW13}: a frequently used similarity method based on the surface matching of mentions and entities; 
\item Word2Vec~\cite{DBLP:journals/bmcbi/ChoCL17}: a new subword embedding is trained on the medical corpus to learn representations.  Cosine similarity is then applied to the average of subword embeddings of each mention-entity pair to rank their closeness;
\item CNN~\cite{DBLP:conf/cnlp/MondalPSGPBG19}: {a CNN-based Siamese network is trained using the triplet loss with the newly trained word2vec embeddings for the  mentions and entities.}
\item BERT~\cite{DBLP:conf/naacl/DevlinCLT19}: the [CLS] representations of mentions and entities are extracted from the fine-tuned BERT to compute their cosine similarity;  
\item DNorm~\cite{DBLP:journals/bioinformatics/LeamanDL13}: {one of the most popular methods that utilizes the TF-IDF embedding and a matching matrix, trained by the margin ranking loss, to determine the similarity score between mentions and entities. }
\item SurfCon~\cite{DBLP:conf/kdd/WangYMHLS19}: {one of the most popular methods that constructs a term-term co-occurrence graph from the raw corpus to capture both the surface information and the global context information for entity synonym discovery.
}
\end{compactenum}

\subsection{Experimental Setup and Evaluation Metrics}%~\cite{DBLP:conf/iclr/KingmaB15}
%\noindent{\bf Experimental setup.} 
The number of sampled negative mention-entity pairs is tuned from \{10, 50, 100, 200, 300\} and set to 200 as it attains the best performance in the development set.  ADAM is adopted as the optimizer with an initial learning rate of 0.001.  The training batch size is 32, and the dimension of the knowledge graph embedding is 200.  Besides, the dimension of the semantic embeddings of both mentions and entities are set to 500, and the dimensions of the first and the second FC networks are set to 300.  These parameters are set by a general value and tuned in a reasonable range.  Dropout is applied in the FC networks and selected as 0.5 from \{0.3, 0.5, 0.7\}.  The knowledge embedding is trained by an open-source package~\footnote{\scriptsize https://github.com/davidlvxin/TransC}.  Early stopping is implemented when the performance in the development set does not improve in the last 10 epochs. % \blue{How about the setup for the baselines?}

%To have a fair comparison, all baseline models share the same setup for the learning rate, the batch size, the embedding dimensions, and the dropout ratio.  For SurfCon~\cite{DBLP:conf/kdd/WangYMHLS19}, we constructed a co-occurrence graph of 24,315 nodes based on the corpus mentioned in the Section \ref{sec:dataset}, and obtained the graph embedding according to the authors~\footnote{https://github.com/yzabc007/SurfCon}.

To provide fair comparisons, we set the same batch size, embedding sizes, and dropout ratio to all baseline models.  For SurfCon, we construct a co-occurrence graph of 24,315 nodes from our collected Chinese medical corpus, and obtain the graph embedding according to~\cite{DBLP:conf/kdd/WangYMHLS19}~\footnote{\scriptsize https://github.com/yzabc007/SurfCon}. 

%\noindent{\bf Evaluation metrics.}  
\emph{Filtered hits@k}, the proportion of correct entities ranked in the top $k$ predictions by filtering out the synonymous entities to the given mention in our constructed $\KG$, because it is an effective metric to determine the accuracy of entity synonyms discovery~\cite{DBLP:conf/emnlp/LvHLL18}.  We follow the standard evaluation procedure~\cite{DBLP:conf/nips/BordesUGWY13,DBLP:conf/emnlp/LvHLL18} and set $k = 3, 5, 10$ to report the model performance. 

%\subsection{Experimental Results}

%\subsubsection{Comparison with baselines}
%\noindent{\bf Experimental results.} 
\subsection{Experimental Results}
Rows three to nine of Table~\ref{tab:results} report the experimental results of the baselines and our \emph{KGSynNet}.  It clearly shows that
\begin{compactitem}[--]
\item JACCARD yields no hit on the difficult case because it cannot build connections on mentions and entities when they do not contain a common sub-word.
\item  {Word2Vec yields the worst performance on the \emph{All} case and the \emph{Regular} case since the representations of mentions and entities are simply obtained by their mean subword embeddings, which blur the effect of each subword.} 
\item CNN improves Word2Vec significantly because of the Siamese network, but cannot even beat JACCARD due to the poor semantic representation learned from Word2Vec. 
\item BERT gains further improvement over JACCARD,  Word2Vec, and CNN by utilizing the pre-trained embeddings.  The improvement is not significant enough especially in the Difficult case because the representation of the token [CLS] does not fully capture the relations between mentions and entities.
\item DNorm further improves the performance by directly modeling the interaction between mentions and entities.  {SurfCon yields the best performance among all baselines because it utilizes external knowledge bases via the term-term co-occurrence graph.}
\item Our \emph{KGSynNet} beats all baselines in all three cases.  Especially, we beat the best baseline, SurfCon, by 14.7\%, 10.3\%, and 5.6\% for the \emph{All} case,
14.2\%, 10.0\%, and 5.4\% for the \emph{Regular} case, and 45.7\%, 24.4\%, and 10.2\% for the \emph{Difficult} case with respect to Hits@3, Hits@5, and Hits@10, respectively.  We have also conducted the statistical significance tests, and observe that for the \emph{All} case group, $p << 0.01$ under the paired t-tests.  The significant improvement clearly shows that our \emph{KGSynNet} is effective in integrating the knowledge information with the semantic features.
\end{compactitem}

%The upper part of Table \ref{tab:results} shows the test results of all compared baseline models. Overall, 

%i) Commonly used methods such as JACCARD and w2v don't perform well in the test set. For the complete cases, their best hits@5 score is less than 58\% and hits@10 less than 64\%. For the difficult cases, they even can't recall any labeled entity in hits@3, which indicates that using only surface information is not sufficient to solve the problem in the difficult cases.

%ii) As both CNN and BERT models have some semantic information incorporated through pretrained embeddings, their results have been improved compared to JACCARD and w2v. For the complete cases, after finetuning on our corpus, BERT reaches 60.41\% of hits@5 and 66.50\% of hits@10. For the difficult cases, both models can correctly predict some of the labeled entities.

%iii) For both regular and difficult cases, DNorm significantly improves the performance compared to all baselines except for SurfCon. For example, for the complete cases, DNorm has a hits@5 value of 63.79\% and a hits@10 value of 71.89\%; for the difficult cases, each hits@k indicator is at least 3 times higher.

%iv) SurfCon has reached the best performance among all baselines on most indicators, illustrating the importance of the term-term co-occurrence graph in the entity alias discovery task.

\begin{table*}[htp]
\centering
\caption{Experimental results: $-$ means that \emph{KGSynNet} removes the component while $\rightarrow$ means that \emph{KGSynNet} replaces the fusion method. }
\label{tab:results}
\scriptsize
\begin{tabular}{@{}l|rrr|rrr|rrr@{}}
\toprule
\multirow{2}{*}{Methods} & \multicolumn{3}{|c|}{hits@3} & \multicolumn{3}{|c|}{hits@5} & \multicolumn{3}{c}{hits@10} \\ \cline{2-10} 
 & \multicolumn{1}{c}{All} & \multicolumn{1}{c}{Regular} & \multicolumn{1}{c|}{Difficult} & \multicolumn{1}{c}{All} & \multicolumn{1}{c}{Regular} & \multicolumn{1}{c|}{Difficult} & \multicolumn{1}{c}{All} & \multicolumn{1}{c}{Regular} & \multicolumn{1}{c}{Difficult} \\ \midrule
JACCARD \cite{DBLP:conf/imecs/NiwattanakulSNW13}  & 52.28\% & 56.61\% & 0.00\% & 58.03\% & 62.83\% & 0.00\% & 63.76\% & 69.04\% & 0.00\%\\
Word2Vec \cite{DBLP:journals/bmcbi/ChoCL17}  & 47.00\% & 50.88\% & 0.00\% & 52.28\% & 56.59\% & 2.30\% & 58.31\% & 63.10\% & 4.60\%\\
CNN \cite{DBLP:conf/cnlp/MondalPSGPBG19}  & 51.76\% & 55.69\% & 4.33\% & 57.75\% & 61.98\% & 6.38\% & 65.13\% & 69.72\% & 9.34\%\\
BERT \cite{DBLP:conf/naacl/DevlinCLT19} & 54.60\% & 58.87\% & 2.96\% & 60.41\% & 65.02\% & 4.78\% & 66.50\% & 71.39\% & 7.52\% \\
DNorm \cite{DBLP:journals/bioinformatics/LeamanDL13} & 56.23\% & 59.78\% & 12.76\% & 63.79\% & 67.58\% & 17.77\% & 71.89\% & 75.64\% & 26.42\% \\
SurfCon \cite{DBLP:conf/kdd/WangYMHLS19}& 58.29\% & 62.02\% & 12.98\% & 66.27\% & 70.11\% & 19.59\%  & 75.20\% & 79.03\% & 28.93\%  \\ \midrule
\emph{KGSynNet} & \textbf{66.84\%} & \textbf{70.81\%} & 18.91\% & \textbf{73.09\%} & \textbf{77.13\%} & 24.37\% & \textbf{79.41\%} & \textbf{83.35\%} & 31.89\% \\%\hdashline 
$-$KE & 64.91\% & 69.07\% & 14.58\% & 71.56\% & 75.77\% & 20.73\% & 79.12\% & 83.14\% & 30.52\% \\
$-$TransC & 65.80\% & 69.92\% & 15.95\% & 71.44\% & 75.79\% & 18.91\%  & 78.94\% & 83.18\% & 27.80\% \\
$\rightarrow$DA & 63.51\% & 67.19\% & \textbf{19.13\%} & 70.85\% & 74.47\% & \textbf{27.10\%} & 78.13\% & 81.77\% & \textbf{34.17\%} \\
$\rightarrow$EF  & 61.98\% & 65.85\% & 15.26\% & 68.63\% & 72.54\% & 21.41\% & 76.28\% & 80.29\% & 27.79\%\\ 
\bottomrule       
\end{tabular}
\end{table*}

%\noindent{\bf Ablation study.}
\subsection{Ablation Study}
\label{sec:ab_study}
{
To better understand why our \emph{KGSynNet} works well, we compare it with four variants: (1) $-$KE: removing the knowledge embedding and the Fusion Gate; (2) $-$TransC: removing losses of Eq.~(\ref{eq:ic_loss}) and Eq.(\ref{eq:cc_loss}) from Eq.~(\ref{eq:TransC-TransE}) of TransC, to learn the knowledge embedding by utilizing only TransE; (3) $\rightarrow$DA: directly adding the learned semantic features and knowledge features of entities together; and (4) $\rightarrow$EF: fusing the learned semantic features and knowledge information via a FC network~\cite{zhang2019ernie}. }

Table~\ref{tab:results} reports the results of the variants in the last four rows and clearly shows three main findings: 
\begin{compactitem}[--]
\item By excluding the knowledge embedding (see the last fourth row in Table~\ref{tab:results}), our \emph{KGSynNet} drops significantly for the All case, i.e., 1.93 for hits@3, 1.53 for hits@3, and 0.29 for hits@10, respectively.  Similar trends appear for the Regular case and the Difficult case.  The performance decay is more serious than those in other variants, $-$TransC and $\rightarrow$DA.  This implies the effectiveness of our \emph{KGSynNet} in utilizing the knowledge information.      
%The external knowledge information plays a significant role in the entity alias prediction. In terms of all metrics, (1st row) performs consistently better than the model without the knowledge embedding (last row). For the complete cases, the difference of their results is much larger in hits@3 (1.93\%) and hits@5(1.53\%) than in hits@10(0.29\%), indicating that our model ranks the correct synonymous entities at a higher place than the knowledge-free model.
\item By removing TransC, we can see that the performance decays accordingly in all cases.  The results make sense because learning the knowledge representation by TransE alone {does not specifically model the \emph{InstanceOf} relation and the \emph{SubclassOf} relation.}  This again demonstrates the effectiveness of our proposed TransC-TransE framework. 

\item In terms of the fusion mechanism, the performance exhibits similarly under the three metrics.  Here, we only detail the results of hits@3.  It shows that the performance by \emph{Fusion Gate} beats ``DA" and ``EF" 3.3 to 5.0 in both the All and Regular cases.  However, ``DA" improves the performance significantly on the Difficult case, i.e., no common sub-word appearing in the mention-entity pairs.  The results make sense because in the Difficult case, the model depends heavily on the external knowledge.  Setting the weight to 1, i.e., the largest weight, on the learned knowledge features can gain more knowledge information.  
On the contrary, ``EF" yields the worst performance on the All and Regular cases, but gains slightly better performance than $-$KE on the Difficult case.  We conjecture one reason is that the available data is not sufficient to trained a more complicated network in ``EF". 
\end{compactitem}

\begin{figure*}[htp]
\centering
  \includegraphics[width=0.8\linewidth]{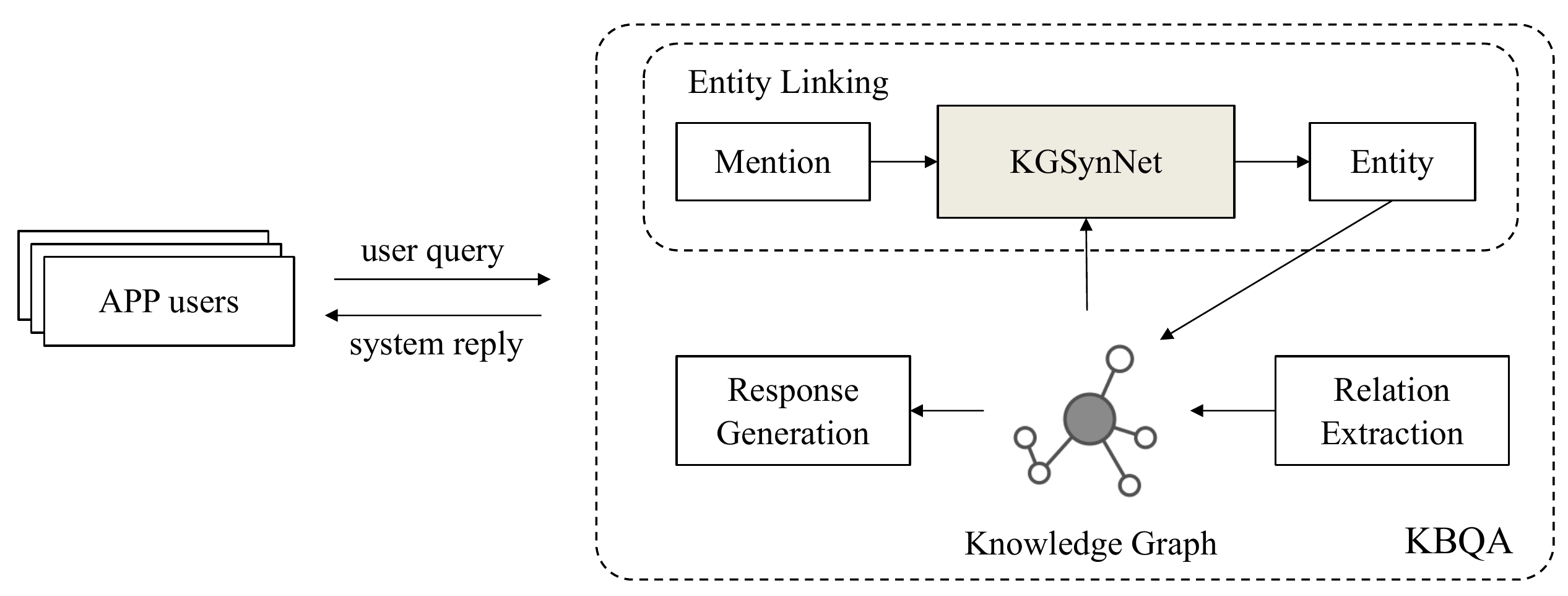}
  \caption{The architecture of online system.}
  \label{fig:online}
\end{figure*}

%\noindent{\bf Online evaluation.}% entity synonyms discovery framework  The existing version utilized BERT model, while the variant deployed our proposed KGSynNet network.   is significant within the confidence level greater than 95\%, which implies that the online performance of KGSynNet is credibly better than the original version using BERT.
\subsection{Online Evaluation}
\label{sec:online_eval} 
Our \emph{KGSynNet} has been deployed in the entity linking module, a key module of the KBQA system of a professional insurance service app, served more than one million insurance agents.  The architecture of the online system is shown in Fig.~\ref{fig:online}.  On average, the requests of the KBQA service of the app are 700K per day with more than 50 requests per second at the peak. 

We conducted an A/B test to compare the original BERT model and our \emph{KGSynNet} on the entity linking module of the KBQA system for two weeks.  The traffic was evenly split into two groups.  Approximately 10\% of users' queries involve disease mentions, within which the proportion of queries with user experience feedback is around 5\%.  Eventually, BERT and \emph{KGSynNet} received about 25K and 26K user feedbacks, respectively. The positive rate of the feedback for BERT is about 34.9\%, while the positive rate of \emph{KGSynNet} is about 37.8\%, significantly better with $p<0.05$ under the paired t-test. 

Moreover, we randomly selected and labeled 1000 disease related queries from each of the two groups. The proportion of queries involving difficult cases was around 3\% in both groups.  Results in Table~\ref{tab:online} show that \emph{KGSynNet} consistently outperforms BERT in terms of hits@3, hits@5, and hits@10, respectively.

\begin{table}[htp]
\centering
\caption{Online evaluation results}
\label{tab:online}
\begin{tabular}{@{}l|rrr|rrr@{}}
\toprule
       \multicolumn{1}{c}{Metric} &
       \multicolumn{3}{|c}{BERT} &
       \multicolumn{3}{|c}{\emph{KGSynNet}} \\ \midrule
 &
  \multicolumn{1}{c}{All} &
  \multicolumn{1}{c}{Regular} &
  \multicolumn{1}{c}{Difficult} &
  \multicolumn{1}{|c}{All} &
  \multicolumn{1}{c}{Regular} &
  \multicolumn{1}{c}{Difficult} \\
hits@3  & 58.2\% & 59.9\% & 3.3\%  & \textbf{68.4\%}   & 70.0\%  & 18.8\%  \\
hits@5  & 63.2\% & 64.9\% & 6.7\%  & \textbf{75.4\%}   & 77.1\%  & 25.0\%  \\
hits@10 & 70.0\% & 71.9\% & 10.0\% & \textbf{81.7\%}   & 83.4\%  & 31.3\%  \\ \bottomrule
\end{tabular}
\end{table}

% One of the key modules of this online KBQA system is entity linking, which strongly depends on our discovered entity aliases for an accurate and high-coverage service. Specifically, we regularly extract new disease aliases from the online logs using our internal NER tool. Then our framework predicts the top-N candidate entities of each alias. After that, these candidate pairs are sent for professional annotations which discover the correct entity-to-alias pairs.

% E行销app拥有超过一百万的保险代理人用户。它搭载的KBQA服务系统，日均访问量70万左右，在高峰时期，每秒并发大于50条。在如此大的访问量下，KBQA服务必须做到快速响应。其次，由于保险相关的咨询涉及法律合规问题，kbqa回答必须非常准确。
% 其中，基于实体别名的实体链接模块是非常重要的一环。所以我们会离线挖掘实体的别名并人工审核，从而保证线上服务响应速度和准确率。
% 基于NER模块，我们定期在更新的日志中抽取疾病别名。然后通过匹配模型，生成别名的Top10候选实体。最后通过人工标注，确定它对应的图谱实体。通过部署KGSynNet，我们发现，相比于原有的基于BERT匹配的模型，标注得到正确实体的比率提高了19.7%。KBQA在线的疾病实体链接准确率达到了98%，提升了用户体验，助力业务增长。

%\if 0
\subsection{Case Studies}\label{sec:cases}
%Here, we provide several typical examples to show the effectiveness of our \emph{KGSynNet}.  In Table~\ref{tab:case1}, we randomly select four query mentions and list the top-5 selected synonymous entities.  The results show that: 
We provide several typical examples to show the effectiveness of our \emph{KGSynNet}.  In Table~\ref{tab:case1}, four query mentions are selected with the top-5 discovered synonymous entities.  The results show that: 
\begin{compactitem}[--]
\item Our \emph{KGSynNet} can successfully detect at least one annotated synonym for each mention.   For example, for the mention, ``hyperelastic skin", our found top-5 synonymous entities are all correct.  
\item For the mention of ``facial paralysis", other than its synonym ``facioplegia", our \emph{KGSynNet} can discover ``prosopoplegia" through the semantic equivalence.  Other top predicted terms, e.g., ``neonatal facial paralysis", ``peripheral facial paralysis", and ``idiopathic facial paralysis", are all hyponyms of the mention with specific clinical manifestations.
\end{compactitem}
%\fi 
%one semantically equal but lexically different synonym "facial paralysis". The other top predicted terms are all hyponyms of it with specific clinical manifestations, which illustrates the effectiveness of our framework. 
% Add examples. Bring back KDD version.

%\if 0
\begin{table*}[htp]
\setlength{\abovecaptionskip}{0.cm}
\setlength{\belowcaptionskip}{-0.cm}
\caption{Query mentions and the corresponding top 5 synonymous entities: the correct synonyms are underlined.}
\label{tab:case1}
\begin{center}
\begin{normalsize}
\resizebox{\textwidth}{!}
{ 
\begin{tabular}{@{}c|l@{}}
\toprule
Mention & Top 5 Found Entities \\ \midrule
弹力过度性皮肤 & \underline{埃莱尔-当洛综合症},\underline{埃勒斯-当洛斯综合症},\underline{皮肤松垂},\underline{埃莱尔-当洛},\underline{皮肤松弛}， \\
hyperelastic skin & \underline{Ehlers-Danlos syndrome}, \underline{Ehlers-Danlos syndrome}, \underline{dermatolysis}, \underline{Ehlers-Danlos}, \underline{cutis laxa} \\\specialrule{0em}{1pt}{1pt} \hline \specialrule{0em}{1pt}{1pt}
肚子痛 & \underline{急性腹泻}, 疼痛, \underline{下腹痛}, 全身疼痛, 疼痛性脂肪过多症 \\
stomachache & \underline{collywobbles}, pain, \underline{hypogastralgia}, generalized pain, lipomatosis dolorosa \\\specialrule{0em}{1pt}{1pt}\hline \specialrule{0em}{1pt}{1pt}\specialrule{0em}{1pt}{1pt}
 & \underline{面瘫}, \underline{面神经麻痹}, 新生儿面部神经麻痹, 周围性面瘫,  \\
歪嘴风 & \underline{prosopoplegia}, \underline{facioplegia}, neonatal facial paralysis, peripheral facial paralysis,  \\\specialrule{0em}{1pt}{1pt}
facial paralysis & 特发性面神经瘫痪\\ 
 & idiopathic facial paralysis\\\specialrule{0em}{1pt}{1pt}
\hline \specialrule{0em}{1pt}{1pt}
倦怠 & \underline{虚弱},\underline{乏力}, 张力失常, 失眠症, 弱精 \\
exhaustion & \underline{debility}, \underline{asthenia}, dystonia, insomnia, asthenozoospermia \\\specialrule{0em}{1pt}{1pt} \bottomrule
\end{tabular}
}
\end{normalsize}
\end{center}
\end{table*}
%\fi 

%\if 0
\subsection{Error Analysis}\label{sec:errors}
We provide a concrete error analysis by sampling 10\% of the incorrectly predicted mention-entity pairs in our \emph{KGSynNet}.  Table~\ref{tab:errortype} lists the main error types: 
\begin{compactitem}[--]
\item More than half of the errors (54\%) occur due to the lack of knowledge in the knowledge graph. {For example, since the entity ``bow legs" is not in the $\KG$, the mention ``knee varus" mistakenly found ``knee valgus" and ``congenital knee valgus" through surface matching.}  %The observation shows that including more external knowledge can further improve the performance. 
% This type of error can be resolved if relevant knowledge was given in the KG, implying the importance of external knowledge in the entity synonym discovery.
\item The second largest error comes from hypernyms distraction, which accounts for 29\% of the total errors. {For example, the mention ``pituitary gland cancer" is distracted to its hypernym ``brain cancer" and ``cerebral cancer", and failed to identify the true entity ``pituitary gland malignant tumor".}
\item Another 12\% of the errors are due to the keyword extraction error.  For example, the golden entity for the mention, ``lung calcification", is ``lung mineralization".  Our \emph{KGSynNet} makes a wrong extraction on the keyword ``calcification" and discovers a wrong entity, ``bronchial calcification", for this mention.  It seems that this problem may be alleviated by adding an fine-grained feature interaction between mentions and entities in our \emph{KGSynNet}.
\end{compactitem}
%\fi 

%\if 0
\begin{table*}[htp]
\setlength{\abovecaptionskip}{0.cm}
\setlength{\belowcaptionskip}{-0.cm}
\caption{Error analysis. The ``Golden Entity" is the correct entity for the corresponding mention.}
\label{tab:errortype}
\begin{center}
\begin{normalsize}
\resizebox{\textwidth}{!}
{ 
\begin{tabular}{p{2.2cm}|l|p{3.2cm}|p{4.5cm}|l}
\toprule
Error Type & Proportion & Mention & Golden Entity & Top 2 Predicted Entities \\ \midrule % \multirow{4}{*}
{Lack of} & \multirow{2}{*}{54\%} & 膝内翻, knee varus & O型腿, bow legs & 膝外翻, knee valgus  \\
Knowledge & &  &  & 先天性膝外翻, congenital knee valgus  \\ 
\hline
 Hypernym Distraction & \multirow{2}{*}{29\%} 
 & 脑垂腺癌, pituitary gland cancer & 垂体恶性肿瘤, pituitary gland malignant tumor & \multirow{2}{*}{\begin{tabular}{l}
 脑癌, brain cancer\\癌性脑病, cerebral cancer\end{tabular}} \\ 
 \hline
{Keyword~Extraction Error} & \multirow{2}{*}{12\%} 
 & 肺部钙化, lung calcification & 肺矿化, lung mineralization & \multirow{2}{*}{\begin{tabular}{l}
 支气管钙化, bronchial calcification \\
 肺转移瘤, pulmonary metastasis   
 \end{tabular}} \\
 \hline
\multirow{2}{*}{Others} & \multirow{2}{*}{5\%}
 & 奥尔布赖特综合症, & 麦凯恩-奥尔布赖特综合症, & 莱特尔综合症, Leiter's syndrome \\
  & & Albright's syndrome & McCune-Albright's syndrome &  吉尔伯特综合症, Gilbert's syndrome \\
  \bottomrule
\end{tabular}
}
\end{normalsize}
\end{center}
\end{table*}
%\fi

\section{Conclusion} 
In this paper, we tackle the task of entity synonyms discovery and propose \emph{KGSynNet} to exploit external knowledge graph and domain-specific corpus.  We resolve the OOV issue and semantic discrepancy in mention-entity pairs.  Moreover, a jointly learned TransC-TransE model is proposed to effectively represent knowledge information while the knowledge information is adaptively absorbed into the semantic features through \emph{fusion gate} mechanism. Extensive experiments and detailed analysis conducted on the dataset show that our model significantly improves the state-of-the-art methods by 14.7\% in terms of the offline hits@3 and outperforms the BERT model by 8.3\% in the online positive feedback rate.  Regarding future work, we can extend our \emph{KGSynNet} to other domains, e.g., education or justice, to verify its generalization ability.  
% Second, we can continually enrich the knowledge graph to improve the discovering ability of our KGSynNet.  Third, we may modify our KGSynNet to tackle the task of entity hypernyms/hyponyms discovery. 

%Future work may include the following: i) extending the framework to other domains with rich external knowledge and highly varied language using styles; ii) continually enriching the KG with professional knowledge; iii) jointly consider the semantic difference between the \emph{synonymous} and the \emph{instanceOf}/\emph{subClassOf} relations. % which takes into account both the semantic meaning and the knowledge information. We use a joint learning model to obtain the knowledge embedding, and have studied how to effectively leverage the knowledge information and incorporate it in the most valuable way. Our solution \emph{Fusion Gate} turns out to be very effective in the integration performance. 
% Meanwhile, \emph{KGSynNet} is capable to discovery the aliases of an entity while simultaneously learning the entity’s comprehensive representation. 
%Taking annotated synonymous disease pairs and a cross-domain knowledge graph as a real dataset, our experiments have demonstrated the effectiveness and efficiency of \emph{KGSynNet}. We have also deployed this framework to one of Ping An's online KBQA systems, achieving an accuracy of 98\% in entity linking. 

\end{CJK*}

\section*{Acknowledgement} 
The authors of this paper were supported by NSFC under grants 62002007 and U20B2053. For any correspondence, please refer to Haiqin Yang and Hao Peng.

\bibliographystyle{splncs04}
\bibliography{KGSynNet}

\begin{thebibliography}{10}
\providecommand{\url}[1]{\texttt{#1}}
\providecommand{\urlprefix}{URL }
\providecommand{\doi}[1]{https://doi.org/#1}

\bibitem{DBLP:journals/ws/BizerLKABCH09}
Bizer, C., Lehmann, J., Kobilarov, G., Auer, S., Becker, C., Cyganiak, R.,
  Hellmann, S.: Dbpedia - {A} crystallization point for the web of data. J. Web
  Semant.  \textbf{7}(3),  154--165 (2009)

\bibitem{DBLP:journals/tacl/BojanowskiGJM17}
Bojanowski, P., Grave, E., Joulin, A., Mikolov, T.: Enriching word vectors with
  subword information. Trans. Assoc. Comput. Linguistics  \textbf{5},  135--146
  (2017)

\bibitem{DBLP:conf/sigmod/BollackerEPST08}
Bollacker, K.D., Evans, C., Paritosh, P., Sturge, T., Taylor, J.: Freebase: a
  collaboratively created graph database for structuring human knowledge. In:
  SIGMOD. pp. 1247--1250. {ACM} (2008)

\bibitem{DBLP:conf/nips/BordesUGWY13}
Bordes, A., Usunier, N., Garc{\'{\i}}a{-}Dur{\'{a}}n, A., Weston, J.,
  Yakhnenko, O.: Translating embeddings for modeling multi-relational data. In:
  NIPS. pp. 2787--2795 (2013)

\bibitem{DBLP:conf/acl/ChenZLWJI17}
Chen, Q., Zhu, X., Ling, Z., Wei, S., Jiang, H., Inkpen, D.: Enhanced {LSTM}
  for natural language inference. In: {ACL}. pp. 1657--1668 (2017)

\bibitem{DBLP:journals/bmcbi/ChoCL17}
Cho, H., Choi, W., Lee, H.: A method for named entity normalization in
  biomedical articles: application to diseases and plants. {BMC} Bioinform.
  \textbf{18}(1),  451:1--12 (2017)

\bibitem{DBLP:conf/naacl/DevlinCLT19}
Devlin, J., Chang, M., Lee, K., Toutanova, K.: {BERT:} pre-training of deep
  bidirectional transformers for language understanding. In: {NAACL}. pp.
  4171--4186 (2019)

\bibitem{DBLP:conf/aaai/DoganL12}
Dogan, R.I., Lu, Z.: An inference method for disease name normalization. In:
  {AAAI} (2012)

\bibitem{DBLP:conf/acl/DSouzaN15}
D'Souza, J., Ng, V.: Sieve-based entity linking for the biomedical domain. In:
  ACL and IJCNLP. pp. 297--302 (2015)

\bibitem{DBLP:conf/naacl/FaruquiDJDHS15}
Faruqui, M., Dodge, J., Jauhar, S.K., Dyer, C., Hovy, E.H., Smith, N.A.:
  Retrofitting word vectors to semantic lexicons. In: {NAACL}. pp. 1606--1615
  (2015)

\bibitem{DBLP:conf/kdd/FeiTL19}
Fei, H., Tan, S., Li, P.: Hierarchical multi-task word embedding learning for
  synonym prediction. In: {ACM} {SIGKDD}. pp. 834--842 (2019)

\bibitem{DBLP:conf/aistats/GutmannH10}
Gutmann, M., Hyv{\"{a}}rinen, A.: Noise-contrastive estimation: {A} new
  estimation principle for unnormalized statistical models. In: {AISTATS}.
  vol.~9, pp. 297--304 (2010)

\bibitem{DBLP:journals/symmetry/HuTZGX19}
Hu, S., Tan, Z., Zeng, W., Ge, B., Xiao, W.: Entity linking via symmetrical
  attention-based neural network and entity structural features. Symmetry
  \textbf{11}(4), ~453 (2019)

\bibitem{DBLP:conf/icdm/JiangLWCLWA13}
Jiang, L., Luo, P., Wang, J., Xiong, Y., Lin, B., Wang, M., An, N.: {GRIAS:} an
  entity-relation graph based framework for discovering entity aliases. In:
  {IEEE} ICDM. pp. 310--319 (2013)

\bibitem{DBLP:journals/bioinformatics/LeamanDL13}
Leaman, R., Dogan, R.I., Lu, Z.: Dnorm: disease name normalization with
  pairwise learning to rank. Bioinform.  \textbf{29}(22),  2909--2917 (2013)

\bibitem{DBLP:journals/bmcbi/LiCTWXWH17}
Li, H., Chen, Q., Tang, B., Wang, X., Xu, H., Wang, B., Huang, D.: Cnn-based
  ranking for biomedical entity normalization. {BMC} Bioinform.
  \textbf{18}({S-11}),  79--86 (2017)

\bibitem{DBLP:conf/emnlp/LvHLL18}
Lv, X., Hou, L., Li, J., Liu, Z.: Differentiating concepts and instances for
  knowledge graph embedding. In: EMNLP. pp. 1971--1979 (2018)

\bibitem{DBLP:conf/nips/MikolovSCCD13}
Mikolov, T., Sutskever, I., Chen, K., Corrado, G.S., Dean, J.: Distributed
  representations of words and phrases and their compositionality. In: NIPS.
  pp. 3111--3119 (2013)

\bibitem{DBLP:conf/cnlp/MondalPSGPBG19}
Mondal, I., Purkayastha, S., Sarkar, S., Goyal, P., Pillai, J., Bhattacharyya,
  A., Gattu, M.: Medical entity linking using triplet network. In: Clinical NLP
  (2019)

\bibitem{DBLP:conf/acl/MouMLXZYJ16}
Mou, L., Men, R., Li, G., Xu, Y., Zhang, L., Yan, R., Jin, Z.: Natural language
  inference by tree-based convolution and heuristic matching. In: ACL (2016)

\bibitem{DBLP:conf/imecs/NiwattanakulSNW13}
Niwattanakul, S., Singthongchai, J., Naenudorn, E., Wanapu, S.: Using of
  jaccard coefficient for keywords similarity. In: IMECS (2013)

\bibitem{DBLP:journals/jamia/SchumacherD19}
Schumacher, E., Dredze, M.: Learning unsupervised contextual representations
  for medical synonym discovery. JAMIA Open  (2019)

\bibitem{DBLP:conf/aaai/ShenLRVSH19}
Shen, J., Lyu, R., Ren, X., Vanni, M., Sadler, B.M., Han, J.: Mining entity
  synonyms with efficient neural set generation. In: {AAAI}. pp. 249--256
  (2019)

\bibitem{DBLP:conf/nips/SrivastavaGS15}
Srivastava, R.K., Greff, K., Schmidhuber, J.: Training very deep networks. In:
  NIPS. pp. 2377--2385 (2015)

\bibitem{DBLP:conf/acl/SungJLK20}
Sung, M., Jeon, H., Lee, J., Kang, J.: Biomedical entity representations with
  synonym marginalization. In: {ACL}. pp. 3641--3650 (2020)

\bibitem{DBLP:conf/ijcai/WangCZ15}
Wang, C., Cao, L., Zhou, B.: Medical synonym extraction with concept space
  models. In: {IJCAI}. pp. 989--995 (2015)

\bibitem{DBLP:conf/edbt/WangLLZ19}
Wang, J., Lin, C., Li, M., Zaniolo, C.: An efficient sliding window approach
  for approximate entity extraction with synonyms. In: {EDBT}. pp. 109--120
  (2019)

\bibitem{DBLP:conf/aaai/WangKMYTACFMMW19}
Wang, X., Kapanipathi, P., Musa, R., Yu, M., Talamadupula, K., Abdelaziz, I.,
  Chang, M., Fokoue, A., Makni, B., Mattei, N., Witbrock, M.: Improving natural
  language inference using external knowledge in the science questions domain.
  In: {AAAI}. pp. 7208--7215 (2019)

\bibitem{DBLP:conf/kdd/WangYMHLS19}
Wang, Z., Yue, X., Moosavinasab, S., Huang, Y., Lin, S.M., Sun, H.: Surfcon:
  Synonym discovery on privacy-aware clinical data. In: {ACM} {SIGKDD}. pp.
  1578--1586 (2019)

\bibitem{zhang2019ernie}
Zhang, Z., Han, X., Liu, Z., Jiang, X., Sun, M., Liu, Q.: {ERNIE:} enhanced
  language representation with informative entities. In: {ACL}. pp. 1441--1451
  (2019)

\end{thebibliography}

\end{document}